\documentclass[sigplan,screen]{acmart}
\usepackage{enumitem}
\usepackage{svg}
\usepackage{algorithm}
\usepackage{algorithmic}
\usepackage{subcaption}

\renewcommand\footnotetextcopyrightpermission[1]{} 
\AtBeginDocument{%
  \providecommand\BibTeX{{%
    \normalfont B\kern-0.5em{\scshape i\kern-0.25em b}\kern-0.8em\TeX}}}

\setcopyright{none}
\begin{document}

\title{Enhancing Conditional Image Generation with Explainable Latent Space Manipulation}

\author{Kshitij Pathania}
\affiliation{%
  \institution{College of Computing, Georgia Institute of Technology}
  \city{Atlanta}
  \state{GA}
  \country{USA}
}
\email{kpathania3@gatech.edu}

\maketitle
\section{Abstract}
In the realm of image synthesis, achieving fidelity to a reference image while adhering to conditional prompts remains a significant challenge. This paper proposes a novel approach that integrates a diffusion model with latent space manipulation and gradient-based selective attention mechanisms to address this issue. Leveraging Grad-SAM (Gradient-based Selective Attention Manipulation), we analyze the cross attention maps of the cross attention layers and gradients for the denoised latent vector, deriving importance scores of elements of denoised latent vector related to the subject of interest. Using this information, we create masks at specific timesteps during denoising to preserve subjects while seamlessly integrating the reference image features. This approach ensures the faithful formation of subjects based on conditional prompts, while concurrently refining the background for a more coherent composition. Our experiments on places365 dataset demonstrate promising results, with our proposed model achieving the lowest mean and median Frechet Inception Distance (FID) scores compared to baseline models, indicating superior fidelity preservation. Furthermore, our model exhibits competitive performance in aligning the generated images with provided textual descriptions, as evidenced by high CLIP scores. These results highlight the effectiveness of our approach in both fidelity preservation and textual context preservation, offering a significant advancement in text-to-image synthesis tasks.

\begin{figure*}
  \centering
  \includegraphics[width=\linewidth]{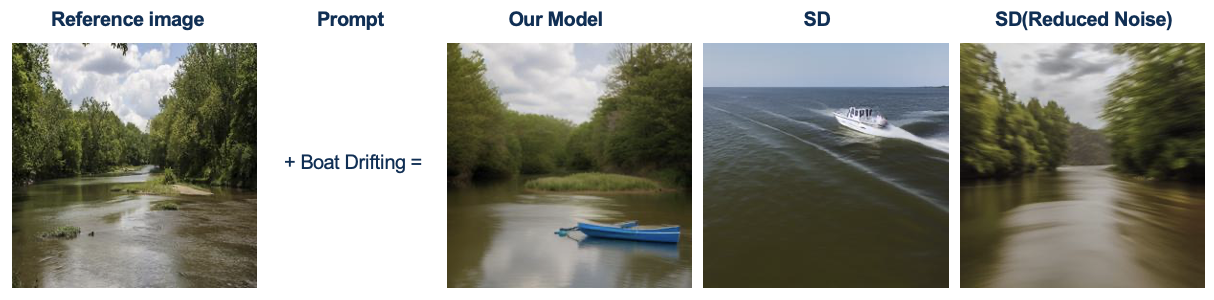}
  \caption{When a reference image is provided alongside a prompt to a diffusion model, SD implementations often struggle to preserve contextual details from the reference. This is primarily due to excessive noise introduction during forward process of DDPM, leading to a loss of contextual fidelity. Conversely, reducing noise can compromise prompt adherence by limiting reconstruction time as evident from the image produced by SD(Reduced Noise). Our proposed model addresses this challenge by maintaining a high number of noise steps while preserving context through targeted replacement of less attended elements in the latent vector. As depicted, our model successfully balances both contextual preservation and prompt adherence.}
  \label{fig:flowchart}
\end{figure*}

\section{Introduction}

Image synthesis, particularly in the context of adhering to both reference images and conditional prompts, poses a significant challenge in contemporary research. While numerous approaches exist, achieving fidelity to reference images while integrating specified conditional prompts remains elusive. This paper presents a novel methodology aimed at addressing this challenge through the integration of a diffusion model with latent space manipulation and gradient-based selective attention mechanisms.

In recent years, diffusion models have emerged as powerful tools for image generation, allowing for the creation of high-quality images by iteratively denoising a latent representation. However, preserving the fidelity of a reference image while adhering to conditional prompts requires further refinement. State-of-the-art models like Stable Diffusion have demonstrated remarkable capabilities in generating images based on a conditional reference image. However, they face challenges when adhering to both the reference image and an additional text condition simultaneously. These models typically operate by controlling the level of noise added to the latent representation, with lesser noise injection required to retain properties from the reference image. This approach often leads to a compromise in faithfully rendering the subjects specified in the text condition, as the reduced number of denoising timesteps limits the model's ability to accurately generate these subjects.

To address this limitation, we propose the integration of Absolute Gradient-based Selective Attention Manipulation for images (Abs-Grad-SAM) motivated from Grad-SAM \cite{barkan2022gradsam}, a technique that analyzes cross-attention maps and gradients within the denoised latent space to generate importance scores. By leveraging Abs-Grad-SAM, we derive importance scores for elements of the denoised latent vector related to the subject of interest. Since we are dealing with images and the Grad-SAM \cite{barkan2022gradsam} technique is originally designed for language models, here, instead of using ReLU, we utilize the absolute values of gradients. This adaptation is crucial as attention can impact pixel values both positively and negatively, requiring a more comprehensive consideration of gradient magnitudes. 

This enables the creation of masks at specific timesteps during denoising, facilitating the preservation of subjects derived from the text condition while seamlessly integrating background elements from the reference image. Consequently, our approach ensures faithful adherence to both the text prompt and the reference image, overcoming the trade-off between subject integrity and background preservation faced by existing methods.

In this paper, we present preliminary findings. Through comprehensive experiments conducted on the Places 365 dataset \cite{bobaaayoung_place365}, we demonstrate the superiority of our proposed method over baseline models. Specifically, our method yields improved Frechet Inception Distance (FID) scores and CLIP scores, indicative of higher-quality image synthesis. Furthermore, our approach generates images that are not only of superior quality but also inherently explainable, accurately reflecting both the specified conditional prompts and the features of reference images. We also address challenges such as object overlap and interference, thus advancing the state-of-the-art in image synthesis techniques.

\section{Related Works}
Denoising Diffusion Probabilistic Models: DDPMs \cite{ho2020denoising} have emerged as a powerful framework for high-quality image synthesis. These models operate by gradually removing noise from an initial noisy latent representation, guided by a neural network trained to predict the noise distribution. [Ho et al., 2020] \cite{ho2020denoising} introduced DDPMs and demonstrated their effectiveness in generating high-quality images on datasets like CIFAR-10 and LSUN. [Nichol and Dhariwal, 2021] \cite{nichol2021improved} extended DDPMs to enable class-conditional image generation, while [Rombach et al., 2022] \cite{rombach2022highresolution} proposed Latent Diffusion Models (LDMs) that apply diffusion models in the latent space of pre-trained autoencoders, enabling high-resolution synthesis with reduced computational requirements.

Text-to-Image Diffusion Models: Building upon the success of DDPMs, researchers have explored techniques to generate images conditioned on text prompts. [Rombach et al., 2022] \cite{rombach2022highresolution} introduced cross-attention layers into the diffusion model architecture, enabling text-to-image synthesis. The Stable Diffusion model [Rombach et al., 2022] \cite{rombach2022highresolution} demonstrated state-of-the-art performance in this domain. However, these models often struggle with faithfully rendering all aspects of the text prompt, leading to the phenomenon of "catastrophic neglect" [Chefer et al., 2023] \cite{chefer2023attendandexcite}. To control the trade-off between sample quality and diversity, classifier guidance [Nichol and Dhariwal, 2021] \cite{nichol2021improved} was introduced, which combines the score estimate of the diffusion model with the estimates of a separately trained image classifier. Alternatively, classifier-free guidance [Ho and Salimans, 2022] \cite{ho2022classifierfree} jointly trains a conditional and an unconditional diffusion model, and combines their respective score estimates to achieve a similar trade-off without requiring a separate classifier. Nichol et al. (2022) \cite{nichol2022glide} introduced GLIDE, a text-guided diffusion model for photorealistic image generation and editing. They compare two guidance strategies—CLIP guidance and classifier-free guidance—and find that the latter is preferred by human evaluators for both photorealism and caption similarity, often producing photorealistic samples. Additionally, they demonstrate the capability of their models for image inpainting, enabling powerful text-driven image editing.

Improving Diffusion Model Performance: Several works have aimed to enhance the performance of diffusion models, particularly in the context of text-to-image generation. [Chefer et al., 2023] \cite{chefer2023attendandexcite} proposed the Attend-and-Excite technique, which guides the model to attend to all subject tokens in the text prompt and strengthen their activations, improving the faithfulness of the generated images. [Hoogeboom and Salimans, 2022] \cite{hoogeboom2022blurring} showed that blurring can be equivalently defined through a Gaussian diffusion process with non-isotropic noise, bridging the gap between inverse heat dissipation and denoising diffusion, and proposed Blurring Diffusion Models that offer the benefits of both standard Gaussian denoising diffusion and inverse heat dissipation. [Lee et al., 2022] \cite{lee2022progressive} introduced a progressive deblurring approach for diffusion models, enabling coarse-to-fine image synthesis by diffusing and deblurring different frequency components of an image at different speeds.. [Ji et al., 2023] \cite{huang2023sat} proposed Self-Attention Control for Diffusion Models Training (SAT), leveraging attention maps to refine intermediate samples during training. [Ruiz et al. 2023] \cite{ruiz2023dreambooth} presented DreamBooth, a method for fine-tuning text-to-image diffusion models to generate personalized images of subjects based on a few reference images. Saharia et al. (2022) \cite{saharia2022photorealistic} introduced Imagen, a text-to-image diffusion model with an unprecedented degree of photorealism and a deep level of language understanding, leveraging techniques from denoising diffusion probabilistic models and latent diffusion models to enhance faithfulness and context adherence in image generation.

In our work, we build upon the advancements in denoising diffusion probabilistic models and latent diffusion models, aiming to improve the faithfulness and context adherence of text-to-image generation. We leverage techniques like attention-based guidance and latent space manipulation to address challenges such as catastrophic neglect and attribute binding, while preserving the strengths of diffusion models in generating high-quality and diverse images.

\section{Methodology}
In our approach, we extend upon the breakthroughs achieved by the state-of-the-art Stable Diffusion model (SD) introduced by Rombach et al. \cite{rombach2022highresolution}  in 2022. Unlike traditional image-based methods, SD operates within the latent space of an autoencoder architecture. Initially, an encoder ($E$) is trained to transform input images ($\mathbf{x} \in \mathbf{X}$) into spatial latent vectors ($\mathbf{z} = E(\mathbf{x})$). We then utilize a denoising diffusion probabilistic model (DDPM) \cite{ho2020denoising} to first add noise to the latent vector through a forward diffusion process, and subsequently denoise the noisy latent vector through a reverse diffusion process over the learned latent space. This process generates a denoised version of the input latent vector $\mathbf{z}$ at each timestep and ultimately we get the final denoised latent vector $\mathbf{\hat{z}}_{d}$. Throughout the denoising procedure, the diffusion model is conditioned on an additional input condition. Following denoising, a decoder ($D$) reconstructs the input image, ensuring that the output of the decoder ($D(\mathbf{\hat{z}}_d)$) closely resembles the desired image ($\mathbf{x}$). Leveraging this latent space framework allows for efficient manipulation and enhancement of image representations, laying the foundation for our integration of importance scores based on Abs-Grad-SAM \cite{barkan2022gradsam} and latent space manipulation techniques.

\begin{figure*}
  \centering
  \includegraphics[width=\linewidth]{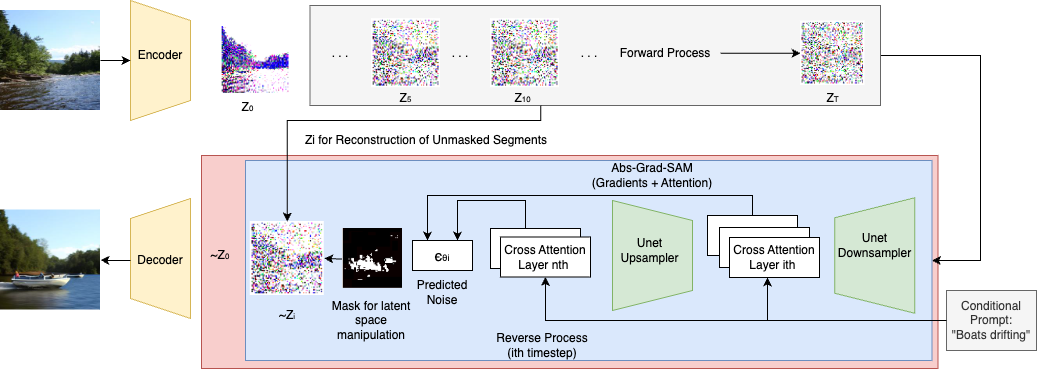}
  \caption{Flowchart illustrating the Abs-Grad-SAM-based latent space manipulation technique for enhancing image generation performance.}
  \label{fig:flowchart}
\end{figure*}

\subsection{Mask Creation}
The denoising diffusion probabilistic model (DDPM) utilizes a UNet architecture to predict noise at each timestamp. Conditional guidance is provided through the use of cross-attention \cite{vaswani2017attention} modules within the UNet. These cross-attention maps within the UNet has varying spatial dimensions and comprises a query, derived from the transformation of the input latent vector, and a key, derived from the input text embedding. Specifically, we define a $\mathbf{n}_{th}$ cross attention map weights as $\mathbf{W}_{A, t, n}$ as a tensor in $\mathbb{R}^{H \times ( P \times P ) \times N}$ where $H$ is the number of attention heads, $P$ is the spatial dimension of the latent vector, and $N$ is the input text size.

For each subject within the input text processed by the cross-attention layer, we compute the Jacobian of gradients of the predicted noise $\mathbf{\hat{z}}_{t}$ with respect to the corresponding attention weights. This calculation is mathematically represented as:
\[
\frac{\partial \mathbf{\hat{z}}_{t}}{\partial \mathbf{W}_{A, t, n}} = \left[ \frac{\partial \mathbf{\hat{z}}_{t}(\mathbf{c}_{i}, \mathbf{x}_{i}, \mathbf{y}_{i})}{\partial \mathbf{W}_{A, t, n}( \mathbf{h}_{j}, \mathbf{s}_{k}, \mathbf{v}_{l})} \right]
\]

where $\mathbf{c}_i$ indexes the channel, $
\mathbf{h}_j$ represent the attention-head, $(\mathbf{x}_i,\mathbf{y}_i)$ represent the spatial coordinates, $\mathbf{s}_k$ indexes the token, and $\mathbf{v}_l$ indexes the variable in the attention map weight vector.

Now, based on the Jacobian calculated and the attention scores for a particular subject $\mathbf{s}_m$, we calculate the importance score of predicted noise elements corresponding to that subject based on the findings from Grad-SAM \cite{barkan2022gradsam}. Mathematically, importance score $\mathbf{I}_{score}$ is given by:
\[
\mathbf{I_{score}(\mathbf{s}_m, \mathbf{c}_{i, t}, \mathbf{x}_{i, t}, \mathbf{y}_{i, t})} = \mathbf{W}_{A, t}(\mathbf{s}_m) \odot \text{Abs}(\mathbf{G}_{\mathbf{s}_m}),
\]

where $\mathbf{G}_{\mathbf{s}_m} := \frac{\partial \mathbf{\hat{z}}_{t}}{\partial \mathbf{W}_{A, t}(\mathbf{s}_m)}$ represents the gradients of the particular subject $\mathbf{s}_m$ in the Jacobian of gradients, and $\odot$ denotes the Hadamard product.

Based on the calculated importance scores, we create a mask $\mathbf{M}_{}(x,y)$ based on a dynamic threshold $\theta$ as described:
\[
\mathbf{M}_{\mathbf{s}_m, n}(c,x,y) = \begin{cases}
1, & \text{if } \mathbf{I}_{score}(\mathbf{s}_m, \mathbf{c}_{i}, \mathbf{x}_{i}, \mathbf{y}_{i}) \geq \theta, \\
0, & \text{otherwise.}
\end{cases}
\]

Additionally, to smooth these masks, we apply a Gaussian filter and morphological dilation on them.
Similarly, we compute the importance score for all the subjects in the input text and generate a mask for each of them. Subsequently, we take union of these masks to create the corresponding mask for the $\mathbf{n}_{th}$ cross-attention layer, which selectively masks all the relevant subjects in the latent vector. Mathematically, we denote this final mask as $\mathbf{M}_{\text{n}}(x,y)$ is given by,

\[
\mathbf{M}_{\text{n}}(c,x,y) = \bigcup_m \mathbf{M}_{\mathbf{s}_m, n}(c,x,y),
\]

Finally, we compute the intersection of these binary masks from all cross-attention layers to derive the ultimate mask $\mathbf{M}_{\text{final}}(x,y)$, serving as the composite representation.:
\[
\mathbf{M}_{\text{final}}(c,x,y) = \bigcap_n \mathbf{M}_n(c,x,y)
\]
\subsection{Latent Space Manipulation}
In latent space manipulation, we manipulate the final denoised latent vector $\mathbf{\hat{z}}_{\text{denoised}}$ for the particular timestamp $t$. We utilize the previously formed mask $\mathbf{M}_{\text{final}}$ to retain the formation of subjects in the image while altering the background from the reference image by replacing the unmasked elements with the corresponding elements from the latent vector $\mathbf{\hat{z}}^{\text{ref}}_t$ obtained by denoising the reference image up to timestep $t$. Mathematically, the manipulated latent vector $\mathbf{\hat{z}}_{\text{manipulated}}$ is given by:
\begin{equation}
\mathbf{\hat{z}}_{\text{manipulated}}(c, x, y) =
\begin{cases}
\mathbf{\hat{z}}_{\text{denoised}}(c, x, y), & \text{if } \mathbf{M}_{\text{final}}(c, x, y) = 1 \\
\mathbf{\hat{z}}^{\text{ref}}_t(c, x, y), & \text{if } \mathbf{M}_{\text{final}}(c, x, y) = 0
\end{cases}
\end{equation}
We perform this latent space manipulation for selected timesteps, which we identify by fine-tuning on a validation set. The choice of these timesteps plays a crucial role in determining the quality of the final output image, as manipulating the latent vector at different timesteps can lead to variations in the preservation of subject details and the incorporation of background elements from the reference image.

\begin{algorithm}
\caption{Latent Space Manipulation for Improved Context Adherence}
\begin{algorithmic}[1]
\REQUIRE Input image $\mathbf{x}$, input text condition, reference image $\mathbf{x}^{\text{ref}}$
\ENSURE Manipulated image $\mathbf{x}_{\text{adhered}}$
\STATE $\mathbf{z} \gets E(\mathbf{x})$ 
\STATE $\mathbf{\hat{z}} \gets \text{DDPM Forward process}(\mathbf{z}, timesteps)$ 

\text{DDPM Reverse process}
\FOR{selected timestep $t$} 
    \STATE $\frac{\partial \mathbf{\hat{z}}_{t}}{\partial \mathbf{W}_{A, t, n}} \gets \text{ComputeJacobian}(\mathbf{\hat{z}}_{t}, \mathbf{W}_{A, t, n})$
    \STATE $\mathbf{I_{score}}(\mathbf{s}_m, \mathbf{c}_i, \mathbf{x}_i, \mathbf{y}_i) \gets \text{CalculateImportanceScores}\left(\frac{\partial \mathbf{\hat{z}}_{t}}{\partial \mathbf{W}_{A, t, n}}, \mathbf{s}_m\right)$
    \STATE $\mathbf{I_{s}} \gets \text{SmoothScores}(\mathbf{I}_{score})$ 
    \STATE $\mathbf{M}_{\mathbf{s}_m, n}(c, x, y) \gets \text{CreateMask}(\mathbf{I_{s}}(\mathbf{s}_m, \mathbf{c}_i, \mathbf{x}_i, \mathbf{y}_i), \theta)$
    \STATE $\mathbf{M}_n(c, x, y) \gets \bigcup_m \mathbf{M}_{\mathbf{s}_m, n}(c, x, y)$ 
    \STATE $\mathbf{M}_{\text{final}}(c, x, y) \gets \bigcap_n \mathbf{M}_n(c, x, y)$ 
    \STATE $\mathbf{\hat{z}}_{\text{ref}}^t \gets \text{DDPM}(E(\mathbf{x}^{\text{ref}}), t)$
    \STATE $\mathbf{\hat{z}}_{\text{manipulated}}(c, x, y) \gets \begin{cases}
    \mathbf{\hat{z}}_{\text{denoised}}(c, x, y), & \text{if } \mathbf{M}_{\text{final}}(c, x, y) = 1 \\
    \mathbf{\hat{z}}_{\text{ref}}^t(c, x, y), & \text{if } \mathbf{M}_{\text{final}}(c, x, y) = 0
    \end{cases}$ 
    \STATE $\mathbf{\hat{z}}_{^{t-1}} \gets \mathbf{\hat{z}}_{\text{manipulated}}$
\ENDFOR
\STATE $\mathbf{\hat{z}}_{denoised} \gets \mathbf{\hat{z}}_{^{0}}$ 
\STATE $\mathbf{x}_{\text{adhered}} \gets D(\mathbf{\hat{z}}_{\text{denoised}})$ 
\end{algorithmic}
\end{algorithm}

\subsection{Mask Refinement and Smoothing}

While the initial masks generated from the importance scores capture the desired regions, their boundaries often exhibit abrupt transitions, which can lead to visual artifacts in the synthesized images. To mitigate this issue and ensure smooth transitions between the preserved subjects and the integrated background elements, we apply two post-processing operations on the importance scores: Gaussian blurring and morphological dilation.

Mathematically, the Gaussian blurring operation can be expressed as:
\[
G(x, y) = \frac{1}{2\pi\sigma^2} \exp\left(-\frac{x^2 + y^2}{2\sigma^2}\right)
\]
\[
I_{\text{blurred}}(x, y) = I(x, y) * G(x, y)
\]
where $I(x, y)$ represents the importance score, $G(x, y)$ is the Gaussian kernel with standard deviation $\sigma$, and $*$ denotes the convolution operation. By applying Gaussian blurring to the generated importance score and use them for generating masks, we effectively smooth the boundaries, reducing abrupt transitions and promoting gradual changes between the preserved and integrated regions.

Morphological dilation is another image processing technique which involves convolving the input mask with a structuring element, typically a small binary kernel. The dilation operation can be mathematically expressed as:
\[
I_{\text{dilated}}(x, y) = \max_{(s, t) \in S} \{I(x - s, y - t) + K(s, t)\}
\]
where $I(x, y)$ is the importance score, $K(s, t)$ is the structuring element (e.g., a disk or square kernel), and $S$ is the domain of the structuring element. By applying dilation to the the scores regions, we slightly expand their boundaries, ensuring a smoother transition between the preserved subjects and the integrated background.

The combination of Gaussian blurring and morphological dilation on the importance scores helps generated masks to mitigate visual artifacts and promote seamless integration of the preserved subjects with the background elements from the reference image. The specific parameters, such as the standard deviation for Gaussian blurring and the structuring element for dilation, are fine-tuned empirically to achieve the desired level of smoothness and boundary preservation.

\section{Experiments and Results}

\subsection{Evaluation Setup}
In our experiments, we utilized the Stable Diffusion v1-5 \cite{rombach2022} model for text-to-image generation tasks. We downloaded the necessary tokenizer files from the Hugging Face model hub and for text embeddings, we employed the CLIP model. This setup allowed us to leverage the capabilities of Stable Diffusion v1-5 for generating high-quality images based on text prompts, while utilizing CLIP for text embeddings to facilitate the generation process.

Additionally, our experiments involved utilizing the Place365 dataset, a subset of which was obtained from kaggle \cite{bobaaayoung_place365}. This dataset consisted of images resized to 256x256 pixels for data compression purposes. We carefully curated the dataset by selecting and identifying specific photos from various categories. The final conditional images dataset had a structured organization, including both artificial and natural scenes each containing several subcategories. Under the artificial category, we have athletic field, industrial area, residential area, downtown, and park. Under the natural category, we have arid area, forest, inland water body, mountain, and ocean. For each subcategory, specific prompts were provided to guide the image generation process. 

To assess the quality of the generated images, we relied on two key metrics: Frechet Inception Distance (FID) \cite{heusel2017gans} scores and CLIP scores \cite{hessel2021clipscore}. FID scores measure the similarity between the distribution of real and generated images in feature space [3], while CLIP scores evaluate the alignment between generated images and corresponding textual prompts [4].

\subsection{Results Analysis}

In this section, we present the results of our experiments on text-to-image generation tasks. Initially, we employed the Stable Diffusion v1-5 model \cite{rombach2022} as our baseline model, coupled with CLIP for text embeddings. To enhance the fidelity of generated images and ensure alignment with the provided conditional image, we fine-tuned the baseline model, identifying optimal hyperparameters. 
This fine-tuned model is referred to as SD tuned. Subsequently, leveraging the dynamic masking technique outlined in the paper, we developed our model. We utilized the Place365 dataset \cite{bobaaayoung_place365} and evaluated the generated images using two key metrics: Frechet Inception Distance (FID) \cite{heusel2017gans} and CLIP scores \cite{hessel2021clipscore}. 

\subsubsection{Frechet Inception Distance (FID) Analysis}

\begin{figure}[htbp]
    \centering
    \includegraphics[width=\linewidth]{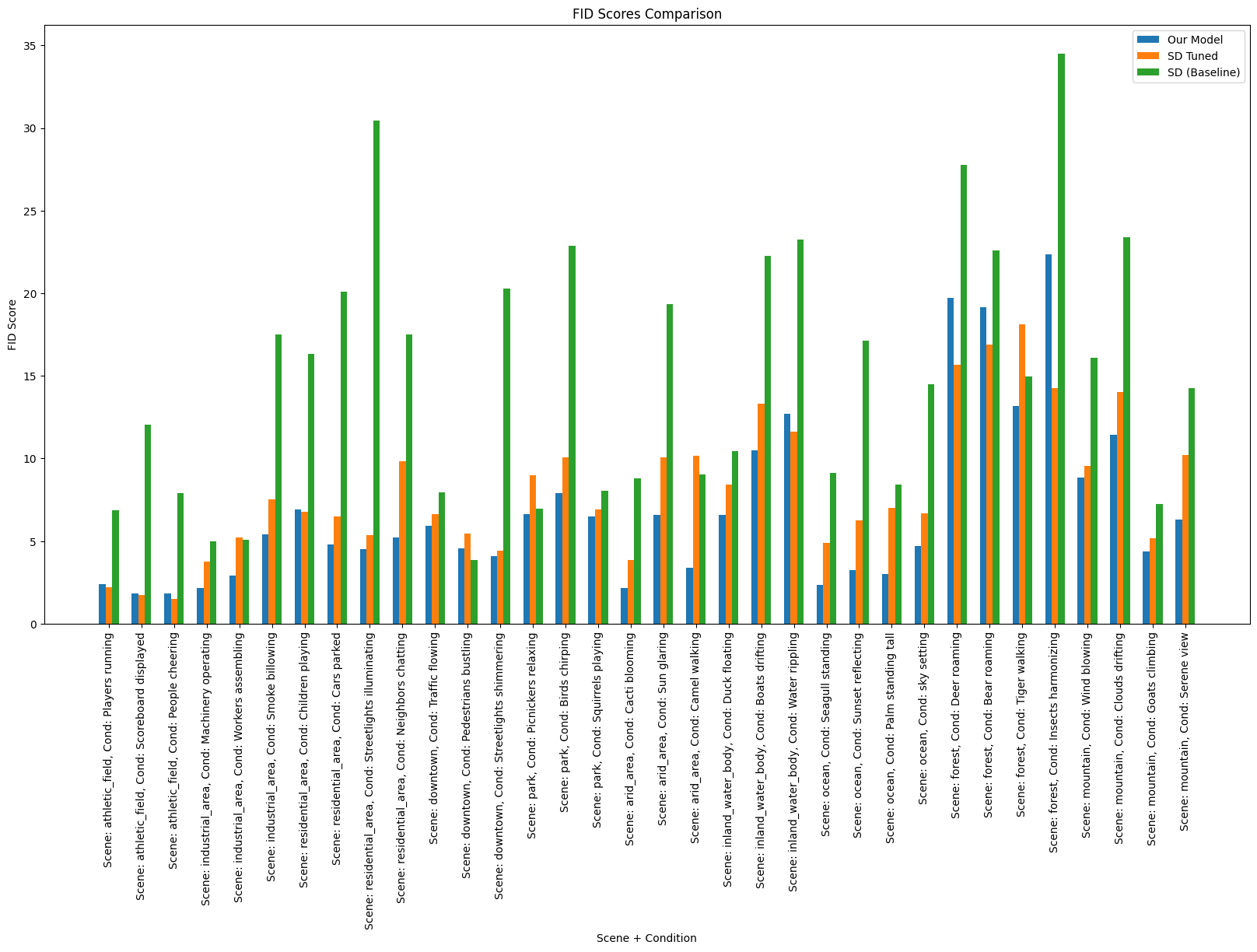}
    \caption{Comparison of FID scores for images generated by different models across different scenes and categories.}
    \label{fig:fid_comparison}
\end{figure}

In the FID scores comparison plot, each bar represents the FID score obtained by a specific model configuration, distinguished by different colors. The x-axis enumerates various scenes and subcategories, providing insights into how each model performs across different visual contexts. Figure~\ref{fig:fid_comparison} shows the FID scores obtained for each model.

\subsubsection{CLIP Score Analysis}
\begin{figure}[htbp]
    \centering
    \includegraphics[width=\linewidth]{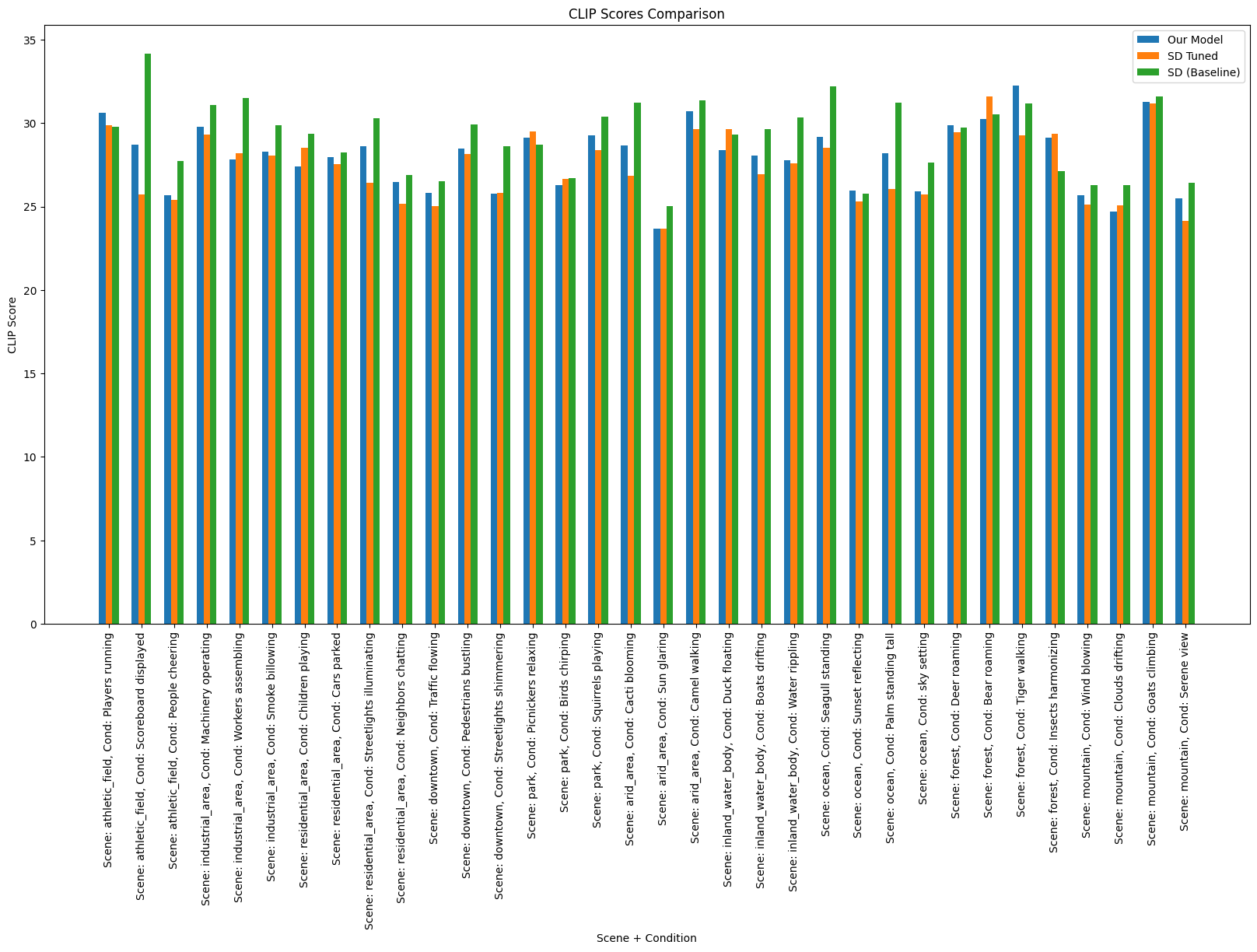}
    \caption{Comparison of CLIP scores for images generated by different models across different scenes and categories..}
    \label{fig:clip_comparison}
\end{figure}

The CLIP scores comparison plot follows a similar format as FID scores, with each bar illustrating the CLIP score achieved by a particular model.
Figure~\ref{fig:clip_comparison} illustrates the CLIP scores obtained for each model.

\subsubsection{Aggregated Results}

\begin{table}[htbp]
    \centering
    \caption{Aggregated Results for FID and CLIP Scores}
    \label{tab:aggregated_results}
    \begin{tabular}{lcc}
        \hline
        \textbf{Model} & \textbf{FID Scores} & \textbf{CLIP Scores} \\
        \hline
        Our Model & \begin{tabular}[c]{@{}c@{}}Mean = 6.89\\ Median = 5.32\end{tabular} & \begin{tabular}[c]{@{}c@{}}Mean = 27.98\\ Median = 28.24\end{tabular} \\
        SD Tuned & \begin{tabular}[c]{@{}c@{}}Mean = 8.21\\ Median = 6.97\end{tabular} & \begin{tabular}[c]{@{}c@{}}Mean = 27.44\\ Median = 27.57\end{tabular} \\
        SD (Baseline) & \begin{tabular}[c]{@{}c@{}}Mean = 15.05\\ Median = 14.72\end{tabular} & \begin{tabular}[c]{@{}c@{}}Mean = 29.19\\ Median = 29.67\end{tabular} \\
        \hline
    \end{tabular}
\end{table}

Our model demonstrates notable performance in both fidelity preservation, as indicated by the Frechet Inception Distance (FID) scores, and textual context preservation, as demonstrated by the CLIP scores. FID scores provide crucial insights into the similarity between the generated images and the real images from the dataset. Lower FID scores imply better fidelity, suggesting that the generated images closely resemble the real images in terms of visual features. Our model achieved the lowest mean and median FID scores compared to the SD Tuned and SD (Baseline) models. With a mean FID score of 6.89 and a median FID score of 5.32, our model demonstrates a high degree of fidelity, maintaining consistency across various scenes and categories. These results indicate our model's effectiveness in capturing the visual characteristics of the scenes from the Place365 dataset, producing images that closely align with real images in terms of visual quality and features. CLIP scores offer insights into the alignment between the generated images and the textual descriptions provided as input. Higher CLIP scores suggest better alignment, indicating that the generated images effectively capture the intended textual context. Despite primarily focusing on fidelity, our model excels in preserving textual context as well. 


With a mean CLIP score of 27.98 and a median CLIP score of 28.24, our model achieves competitive performance in aligning the generated images with the provided textual descriptions. This highlights our model's capability to not only produce visually appealing images but also ensure that these images accurately reflect the intended textual context. Overall, our model's balanced performance in both fidelity preservation and textual context preservation demonstrates its effectiveness in text-to-image generation tasks, providing high-quality image generation while maintaining fidelity to the provided textual descriptions.

\subsection{Qualitative Results}

\begin{figure}
    \centering
    \begin{subfigure}{0.45\textwidth}
        \centering
        \includegraphics[width=\linewidth]{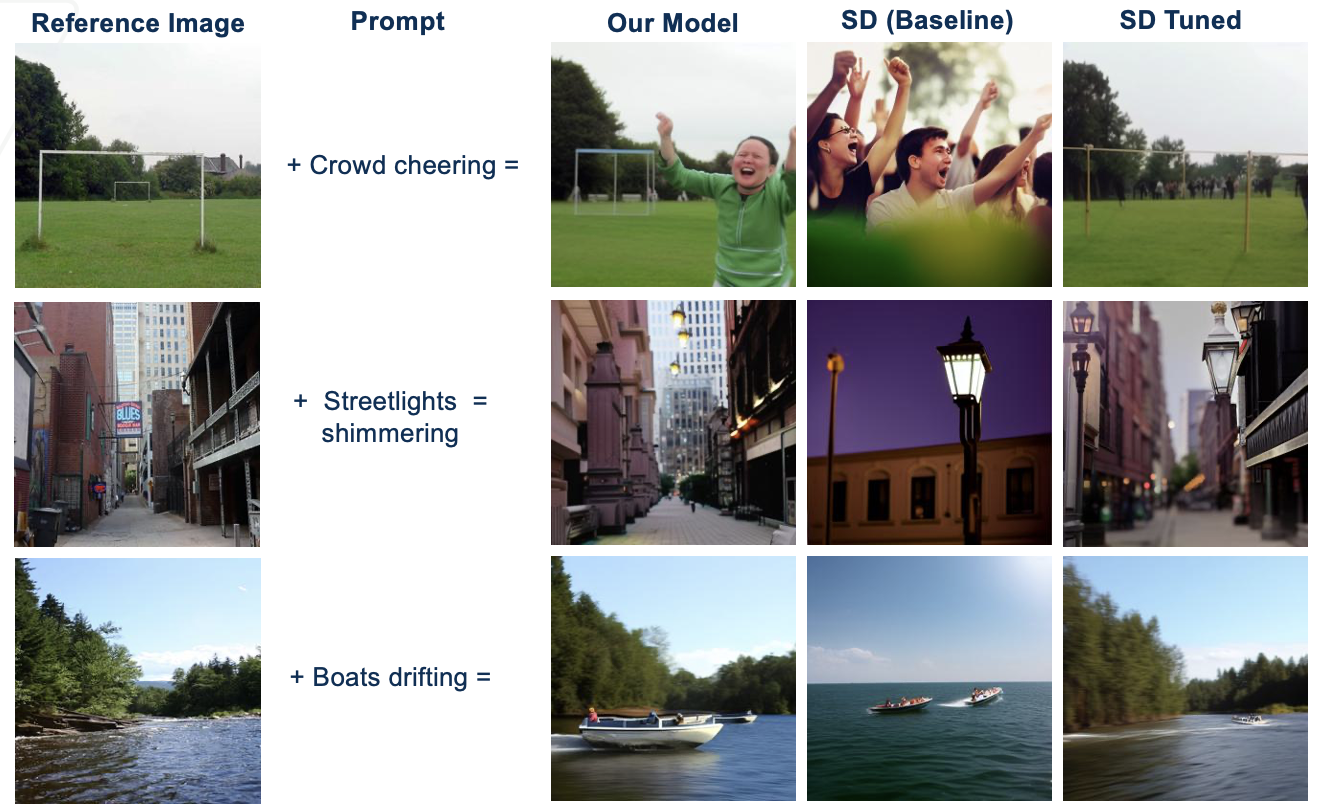}
        \caption{}
        \label{fig:sub1}
    \end{subfigure}%
    \hspace{1\textwidth}
    \begin{subfigure}{0.44\textwidth}
        \centering
        \includegraphics[width=\linewidth]{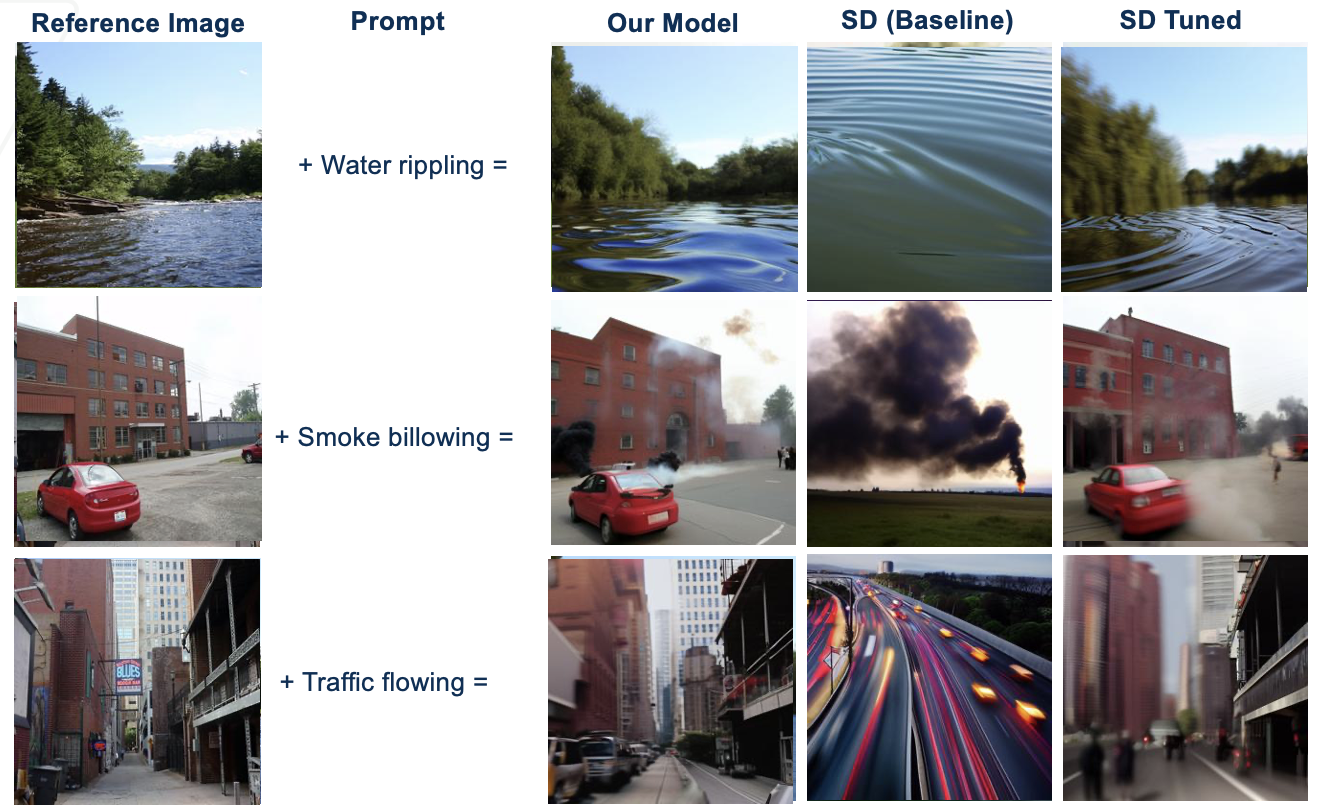}
        \caption{}
        \label{fig:sub2}
    \end{subfigure}
    \caption{A Comparative analysis.}
    \label{fig:flowchart}
\end{figure}

Based on qualitative analysis of the images presented in comparative analysis, it is apparent that our model exhibits a notably higher adherence to the specified condition in comparison to the tuned model. While the baseline model strictly adheres to the condition, it falls short in integrating essential features and contextual cues from the reference image, primarily attributable to the increased noise present in its latent vector.

 Conversely, the tuned model displays adeptness in capturing features from the reference image, facilitated by a latent vector having reduced noise levels. However, this diminished noise level contributes to a compromised fidelity in incorporating conditional elements into the image. Our proposed model effectively preserves the contextual integrity of the image while concurrently upholding fidelity to the prescribed condition.
 
\section{Conclusion}

In this study, we explored text-to-image generation tasks using the Stable Diffusion model coupled with Abs-Grad-SAM technique for latent space manipulation. Through experimentation, we fine-tuned the baseline model and developed our model by leveraging the dynamic masking technique. We evaluated the generated images using Frechet Inception Distance (FID) scores and CLIP scores, which provided insights into fidelity preservation and textual context alignment, respectively.

Our model exhibited superior performance compared to both the baseline SD model and the fine-tuned SD model (SD Tuned). With lower FID scores indicating better fidelity and competitive CLIP scores suggesting effective alignment with textual prompts, our model demonstrated balanced performance in both visual quality and textual context preservation.

Furthermore, the incorporation of the Abs-Grad-SAM method enabled us to manipulate latent space vectors in an explainable manner, potentially enhancing image precision and control over the generation process. While our experiments utilized importance scores from a single cross-attention layer, future research could explore leveraging all cross-attention layers for even more precise image generation.

Overall, our findings underscore the effectiveness of our model in text-to-image generation tasks, offering high-quality image generation while maintaining fidelity to textual descriptions. This research contributes to advancing the capabilities of generative models and opens avenues for further exploration in controlled image generation and explainable latent space manipulation.

\section{Acknowledgements}

This research was conducted as part of the CS 7476 Advanced Computer Vision course in Spring 2024. I thank Professor James Hays and the course Teaching Assistant Akshay Krishnan for their guidance and support throughout the project.

\bibliographystyle{ACM-Reference-Format}
\bibliography{sample-base}

\end{document}